\documentclass[letterpaper]{article} % DO NOT CHANGE THIS
\usepackage{aaai25}  % DO NOT CHANGE THIS
\usepackage{times}  % DO NOT CHANGE THIS
\usepackage{helvet}  % DO NOT CHANGE THIS
\usepackage{courier}  % DO NOT CHANGE THIS
\usepackage[hyphens]{url}  % DO NOT CHANGE THIS
\usepackage{graphicx} % DO NOT CHANGE THIS
\usepackage{color}
\usepackage{amsfonts,amssymb}
\usepackage{amsmath} 
\usepackage{natbib}  % DO NOT CHANGE THIS AND DO NOT ADD ANY OPTIONS TO IT
\usepackage{caption} % DO NOT CHANGE THIS AND DO NOT ADD ANY OPTIONS TO IT
\usepackage{algorithm}
\usepackage{algorithmic}
\usepackage{newfloat}
\usepackage{listings}
\DeclareCaptionStyle{ruled}{labelfont=normalfont,labelsep=colon,strut=off} % DO NOT CHANGE THIS
\floatstyle{ruled}
\newfloat{listing}{tb}{lst}{}
\floatname{listing}{Listing}
\title{RealisHuman: A Two-Stage Approach for Refining Malformed Human Parts in Generated Images}
\author{
    Benzhi Wang \textsuperscript{\rm 1,\rm 2}\equalcontrib,
    Jingkai Zhou \textsuperscript{\rm 4}\equalcontrib, 
    Jingqi Bai \textsuperscript{\rm 1, \rm 2}\equalcontrib, \\
    Yang Yang \textsuperscript{\rm 1, \rm 2},
    Weihua Chen \textsuperscript{\rm 4}\textsuperscript{\textdagger},
    Fan Wang \textsuperscript{\rm 4},
    Zhen Lei \textsuperscript{\rm 1, \rm 2, \rm 3}\textsuperscript{\textdagger}
}

\affiliations{
\textsuperscript{\rm 1} State Key Laboratory of Multimodal Artificial Intelligence Systems, \\ Institute of Automation, Chinese Academy of Sciences\\
\textsuperscript{\rm 2}School of Artificial Intelligence, University of Chinese Academy of Sciences\\
\textsuperscript{\rm 3}Centre for Artificial Intelligence and Robotics, Hong Kong Institute of Science \& Innovation,\\
Chinese Academy of Sciences\\
\textsuperscript{\rm 4} Alibaba Group

    wangbzzz@foxmail.com, \{yang.yang, zlei\}@nlpr.ia.ac.cn, baijingqi2022@ia.ac.cn \\ \{zhoujingkai.zjk, kugang.cwh, fan.w\}@alibaba-inc.com
}

\usepackage{bibentry}
\nocopyright
\begin{document}

\maketitle
\renewcommand{\thefootnote}{\fnsymbol{footnote}}
\footnotetext[\value{footnote}]{\textdagger \ Corresponding Authors}
\begin{abstract}
In recent years, diffusion models have revolutionized visual generation, outperforming traditional frameworks like Generative Adversarial Networks (GANs). 
However, generating images of humans with realistic semantic parts, such as hands and faces, remains a significant challenge due to their intricate structural complexity. To address this issue, we propose a novel post-processing solution named RealisHuman. The RealisHuman framework operates in two stages. First, it generates realistic human parts, such as hands or faces, using the original malformed parts as references, ensuring consistent details with the original image. Second, it seamlessly integrates the rectified human parts back into their corresponding positions by repainting the surrounding areas to ensure smooth and realistic blending. The RealisHuman framework significantly enhances the realism of human generation, as demonstrated by notable improvements in both qualitative and quantitative metrics. Code is available at \textcolor{blue}{\url{https://github.com/Wangbenzhi/RealisHuman}}.
\end{abstract}

\section{Introduction}
Diffusion models have emerged as a powerful approach in the field of visual generation, significantly surpassing traditional frameworks such as Generative Adversarial Networks (GANs) \cite{gans}. These models function as parameterized Markov chains, showcasing an exceptional capability to convert random noise into complex images through a sequential refinement process. Starting with noise, diffusion models progressively enhance the visual quality, ultimately producing high-fidelity representations. With ongoing technological advancements, diffusion models have shown substantial promise in image generation and various related tasks\cite{sdxl,stable_diffusion,animatediff}.

\begin{figure}[h!]
    \centering
\includegraphics[width=\columnwidth]{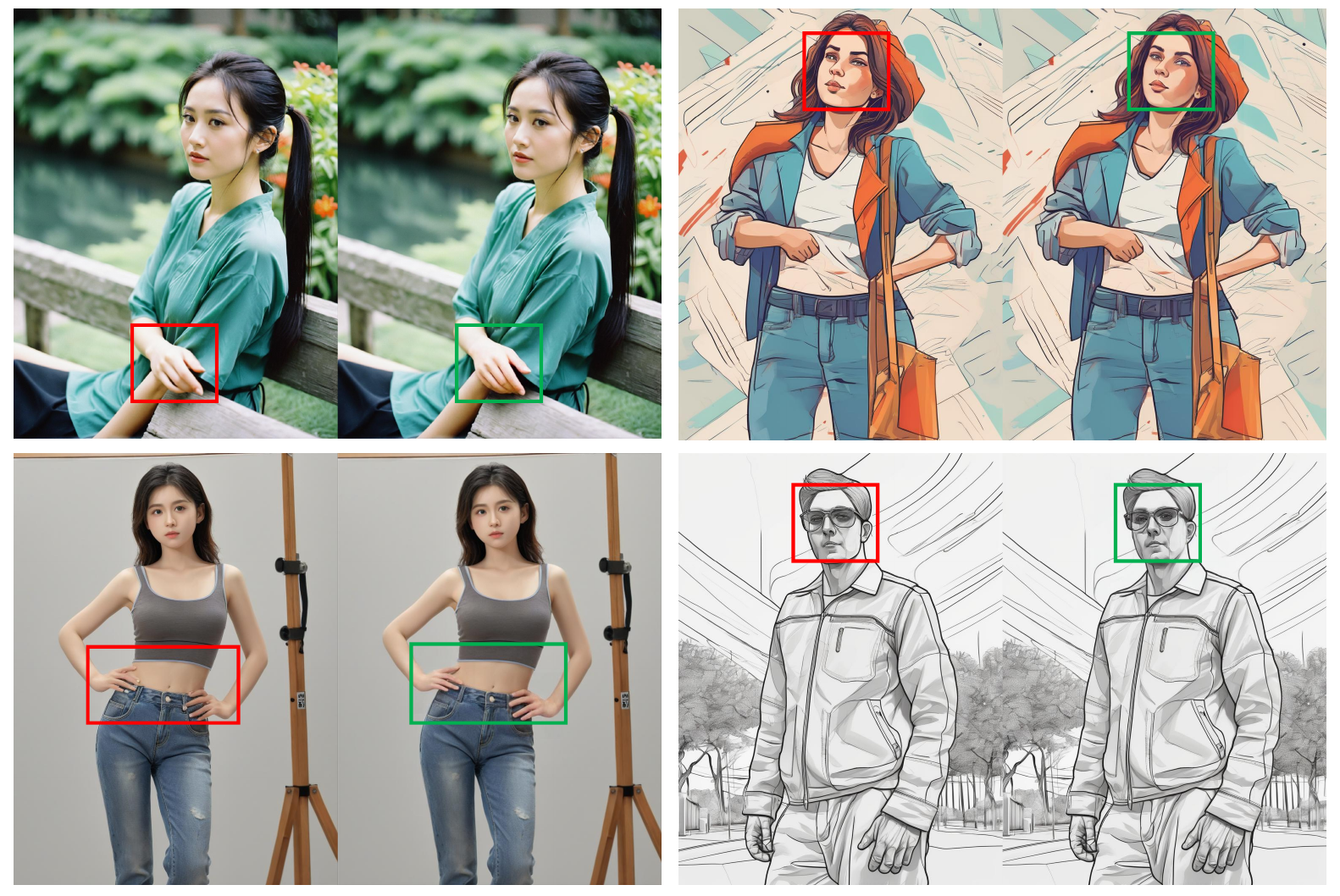}
\caption{Illustration of our repair results. Each pair consists of two images: the left image is the original, and the right image is the repair result. }
\label{fig:fig1}
\end{figure}
Despite their remarkable performance in generating a diverse range of objects, diffusion-based models encounter significant challenges when reconstructing realistic human features, particularly faces and hands.  The intricate structural complexity of these parts, coupled with the limited information preserved after VAE encoder downsampling \cite{auto-encoder}, often leads to incorrect hand structures or distorted faces.
As depicted in Fig.\ref{fig:fig1}, these inaccuracies highlight the difficulties these models face in human image generation.

To address this issue, HandRefiner\cite{lu2023handrefiner} proposed a lightweight post-processing solution that employs a conditional inpainting approach to correct malformed hands while preserving other image regions. Utilizing a hand mesh reconstruction model, HandRefiner ensures accurate finger counts and hand shapes, fitting the desired hand pose. By leveraging ControlNet modules, HandRefiner reintegrates correct hand information into the generated images, enhancing overall image quality. However, this method has two notable limitations. As illustrated in Fig.\ref{fig:compare-with-handrefiner}, HandRefiner often fails to maintain consistency in skin tone and texture due to missing reference information. It also struggles with reconstructing detailed hands when the regions are small. Additionally, it can introduce distortions in other areas, like the face, compromising the overall image integrity.

\begin{figure*}[h!]
\centering
\includegraphics[width=2\columnwidth]{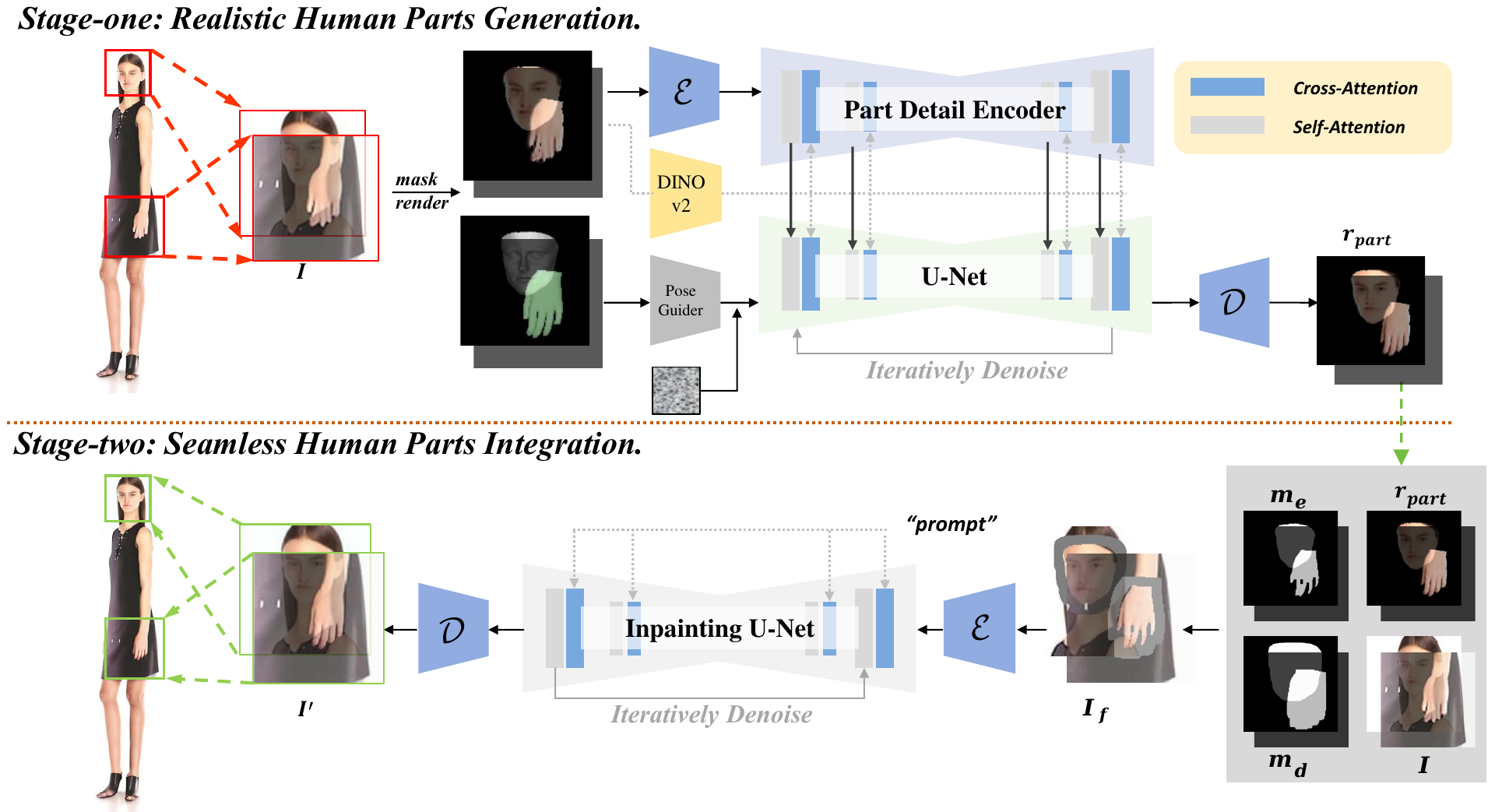} 
\caption{Details of our RealisHuman. Our method separates the task of refining malformed human parts into two distinct stages. In the first stage, we focus on generating realistic human parts using the Part Detail Encoder. Given an image containing malformed human parts, we begin by locating and cropping the target regions. Subsequently, we filter the background of the target regions, creating a reference image that provides essential part details, such as skin tone. We also estimate the 3D structure of the human parts to serve as pose guidance. Leveraging both the reference images and the part structures, we generate realistic human parts $r_{part}$ with accurate structures and detailed information. In the second stage, our goal is to seamlessly integrate the refined human parts into the corresponding regions of the original image, resulting in the refined image $I^{'}$. To avoid a cut-and-paste appearance, we repaint the area between the background and the rectified human parts, ensuring a seamless integration and a more natural overall appearance.}
\label{fig:framework}
\end{figure*}

In this paper, we propose a novel post-processing solution named RealisHuman to address the challenge of refining malformed human parts. To ensure high-quality refinements in small regions, our method locates and crops the malformed areas, allowing us to concentrate on detailed local refinements. Compared to HandRefiner, our method is capable of refining various human parts, not just hands, while preserving intricate details such as skin tone and texture. This capability ensures that the refined parts are both realistic and consistent with the surrounding image. Additionally, our approach demonstrates strong generalization capabilities, effectively handling different styles of images, including cartoons, sketches, and so on. As shown in Fig.\ref{fig:framework}, our RealisHuman framework operates in two stages. In the first stage, our goal is to generate rectified human parts that preserve the consistent details of the original malformed parts. By using the malformed parts as references, we extract detailed information through the Part Detail Encoder and DINOv2, ensuring the preservation of fine-grained details and enhancing the overall realism of the generated parts. Additionally, we incorporate 3D pose estimation results extracted from the malformed parts to guide the generation of human part images, ensuring that the poses are both accurate and realistic.  After obtaining the rectified human parts, the subsequent challenge is to seamlessly integrate them into the original local image. We address this as an inpainting problem. Initially, the rectified human parts are placed back into their original positions, and the surrounding areas are masked. We then train a model capable of seamlessly blending the human parts with the surrounding areas, ensuring a smooth transition and realistic integration. Finally, the refined human parts are pasted into the original image, completing the process of malformed human parts refinement. This approach not only corrects structural inaccuracies but also maintains visual coherence with the original image, providing a robust solution for human parts refinement in image generation tasks. The RealisHuman framework significantly enhances the realism of human generation, as validated by comprehensive experiments demonstrating improvements in both qualitative and quantitative measures.

Our contributions are summarized as follows:

\begin{itemize}
    \item We propose a novel post-processing framework named RealisHuman to address the task of refining human parts in generated images. Our method maintains consistent details with the original image, effectively handles small part refinements, and demonstrates strong generalization across different image styles.
    
    \item We propose a novel two-stage local refinement paradigm, which can be extended to the refinement of other structurally fixed objects, such as distorted logos.
    %consists with generating realistic human parts and seamlessly integrating them into the original image.
    
    \item The RealisHuman framework significantly enhances the realism of human generation, as evidenced by extensive experiments demonstrating enhancements in both qualitative and quantitative metrics.

\end{itemize}

\section{Related Work}
\noindent\textbf{Diffusion Model for Image Generation.} 
Recently, diffusion models have attracted a lot of attention because of their powerful generating ability and have become a hot research direction in the field of computer vision. These models have exhibited superior performance, surpassing conventional techniques due to their intrinsic capability to generate high-quality and diverse outputs. However, the high dimensionality of images introduces significant computational complexity. To address this, the Latent Diffusion Model (LDM) \cite{rombach2022high} was proposed. LDM performs denoising within a lower-dimensional latent space using a pre-trained autoencoder. This approach effectively balances computational efficiency with generative performance, representing a pivotal advancement in the scalability of diffusion-based image generation.
Despite these advancements, controlling the generative process of diffusion models remains a challenge, particularly when precise semantic adherence is required. Diffusion models have achieved great success in producing realistic images that adhere to the semantic content provided by encoding text inputs into latent vectors via pre-trained language models like CLIP \cite{CLIP}. However, relying solely on text descriptions for controlling the model is insufficient, especially when it comes to describing postures and actions \cite{ye2023affordance}. 
To enhance controllability and precision in generated imagery, researchers have explored the incorporation of additional control signals. ControlNet \cite{zhang2023adding} employs a trainable duplicate of the Stable Diffusion (SD) encoder architecture to extract features from conditional inputs. Similarly, T2I-Adapter \cite{mou2024t2i} utilizes lightweight, composable adapter blocks for feature extraction. These additional conditional layers have proven instrumental in improving the model's controllability under various conditions, such as pose, mask, and edge, thereby significantly influencing the direction of its output.

\noindent\textbf{Realistic Human Image Generation.}
Diffusion models have been extensively utilized for pose-conditioned human image synthesis tasks. Animate Anyone \cite{animateanyone} proposes a novel network architecture, ReferenceNet, specifically designed as a symmetrical UNet structure to capture the spatial details of reference images. MagicAnimate \cite{xu2024magicanimate} adopts a similar approach but utilizes a ControlNet specifically tailored for DensePose \cite{guler2018densepose} inputs instead of the more commonly used OpenPose \cite{openpose} keypoints, thereby offering more precise pose guidance. Champ \cite{zhu2024champ} incorporates four distinct control signals simultaneously as conditions for guiding the image generation process, namely depth, normal, semantic, and skeleton, which are extracted from SMPL \cite{loper2023smpl} models.
Despite the remarkable advancements in generating high-quality synthetic images of humans, a persistent challenge remains in the synthesis of hands. This is primarily due to the intricate nature of hand anatomy and the difficulty in accurately depicting hands using skeletal frameworks. Some approaches have begun to specifically focus on generating higher-quality hands. Diffusion-HPC \cite{weng2023diffusion} introduces a technique that employs depth maps of human bodies rendered from reconstructed human body meshes, utilizing conditional diffusion models to correct morphological abnormalities in generated human bodies. Similarly, HandRefiner \cite{lu2023handrefiner} proposes a post-processing approach that utilizes a reconstructed hand mesh to provide essential information about hand shape and location. While these methods can be effective in addressing distortions in hand morphology, they often fall short in preserving fine details such as skin tone consistency and texture.

To enhance the realism of human generation, we propose a two-stage post-processing method named RealisHuman. In the first stage, our method rectifies malformed parts by utilizing detailed information and 3D pose estimation results from the original malformed parts. In the second stage, we seamlessly integrate the rectified human parts back into the original image to complete the refinement process.

\section{Method}
Our goal is to refine the malformed parts while preserving the consistent details of the original parts. The overall framework pipeline is depicted in Fig.\ref{fig:framework}. To ensure the realism of the rectified human  parts, the pipeline is divided into two distinct stages. In the first stage, the rectified human parts are generated under the guidance of the parts meshes and the malformed part images. In the second stage, the rectified human parts obtained from the first stage are integrated back into the local image, followed by repainting the surrounding region to achieve the final results.
\subsection{Preliminary}
\textbf{Latent Diffusion Models.} Our approach builds upon the foundation of Stable Diffusion (SD)\cite{stable_diffusion}, which originates from the Latent Diffusion Model (LDM). LDMs are designed to operate within the latent space managed by an autoencoder, specifically $\mathcal{D}(\mathcal{E}(\cdot))$. A prime example of these models is Stable Diffusion (SD), which combines a Variational AutoEncoder (VAE)\cite{auto-encoder} and a time-conditioned U-Net\cite{unet} to estimate noise. For handling text inputs, SD uses a CLIP ViT-L/14\cite{CLIP} text encoder to transform textual queries into embeddings, denoted as $c_{\text{text}}$.

In the training phase, the model processes an image $I$ and a corresponding text condition $c_{\text{text}}$. The image is encoded into a latent representation $z_0 = \mathcal{E}(I)$, which then undergoes a predefined sequence of $T$ diffusion steps governed by a Gaussian process, resulting in a noisy latent representation $z_T \sim \mathcal{N}(0, 1)$. The objective of SD is to iteratively refine $z_T$ back to $z_0$, using the following loss function:

\begin{equation}
    L = \mathbb{E}_{E(I), c_{\text{text}}, \epsilon \sim \mathcal{N}(0,1), t} \left[ \left\| \epsilon - \epsilon_\theta(z_t, t, c_{\text{text}}) \right\|_2^2 \right],
\end{equation}
where $t = 1, ..., T$ denotes the timestep embedding. $\epsilon_\theta$ denotes the trainable components within the denoising U-Net, which processes the noisy latents $z_t$ and the text condition $c_{\text{text}}$. The architecture of the U-Net includes convolutional layers (Residual Blocks) and both self- attention and cross-attention mechanisms (Transformer Blocks).

The training process involves encoding the image into a latent form $z_0$ and subjecting it to a sequence of diffusion steps, producing $z_T$. The denoising U-Net is trained to predict and remove the noise added during these steps. Once trained, the model can generate $z_0$ from $z_T$ using a deterministic sampling method (such as DDIM\cite{DDIM}), and the final image is reconstructed through the decoder $\mathcal{D}$.

During inference, the initial latent $z_T$ is sampled from a Gaussian distribution with the initial timestep T and gradually refined through iterative denoising steps to yield $z_0$. At each step, the U-Net predicts the noise present in the latent features corresponding to that specific timestep. The decoder $\mathcal{D}$ then reconstructs the final image from $z_0$.

\subsection{Realistic Human Parts Generation}\label{stage1}
In the first stage, our objective is to generate realistic parts that maintain consistent detail and pose with the original images. This is achieved by using the guidance of meshes and reference information from the malformed parts. Leveraging these, we ensure the rectified human parts match the intended appearance and pose.

\noindent\textbf{Data preparation.} Suppose we have a series of original human images and corresponding generated images that contain malformed human parts, produced by algorithms such as \cite{stable_diffusion, zhu2024champ, animateanyone, wang2023disco, chang2024magicpose, xu2024magicanimate, karras2023dreampose}. We begin by locating and cropping the target part regions using the human skeleton estimation method \cite{dwpose}. After isolating the target part regions, we employ the state-of-the-art (SOTA) mesh reconstruction method \cite{hamer, 3ddfa_v3} to estimate the meshes for each part. Additionally, we render the meshes to produce depth maps and binary mask maps $m$. To reduce the influence of the background and focus on realistic human parts generation, we apply the mask $m$ to filter out the background and obtain the foreground regions of the human parts as reference images $I_{ref}$.

\noindent\textbf{Part Detail Encoder.} Previous image-conditioned generation tasks \cite{wang2023disco,karras2023dreampose} have typically utilized the CLIP image encoder \cite{CLIP} to encode reference images. Specifically, these methods compress reference images from a spatial size of 224 $\times$ 224 $\times$ 3 into a one-dimensional vector of dimension 1024, and then employ cross-attention mechanisms to integrate the latent representation with this vector. However, these approaches face challenges in preserving appearance details, as encoding reference images into semantic-level features results in a loss of spatial representations. Previous works \cite{animateanyone, cao2023masactrl, chang2024magicpose} have demonstrated that the self-attention mechanism can significantly enhance the preservation of detail in reference images. Inspired by these findings, we introduce the Part Detail Encoder to improve the realism of rectified human parts by integrating detailed information from the reference images $I_{ref}$. The Part Detail Encoder shares the same architecture as the original Stable Diffusion (SD), comprising self-attention and cross-attention layers, and is initialized with the original SD UNet. To achieve this, we use the reference images as input to the Part Detail Encoder and obtain intermediate outputs. To better integrate detailed information, we modify the input to the self-attention mechanism of the UNet. Specifically, we concatenate the intermediate outputs of the Part Detail Encoder with those of the original SD, and use this concatenated output as the input to the self-attention mechanism of the original SD. This approach ensures that fine-grained details are preserved, enhancing the overall realism of the generated human parts. The modified self-attention mechanism can be formulated as:
\begin{equation}
\small    \text{$f_{s}$}=\text{softmax}\left(\frac{Q_o \cdot (K_o \oplus K_h)^T}{\sqrt{d}}\right) \cdot (V_o \oplus V_h),
\end{equation}
where $d$ is the feature dimension. $Q_o$, $K_o$, and $V_o$ denote the query, key, and value from the self-attention layers of the original SD, respectively. Meanwhile, $K_h$ and $V_h$ denote the key and value from the self-attention layers of the Part Detail Encoder. 

Meanwhile, we employ DINOv2\cite{oquab2023dinov2} to get the image embedding $c_r$ of the reference image, which is then passed into the model through a cross-attention mechanism. This approach supplements the semantic-level features of the reference image. The depth map is processed through several convolution layers to obtain the pose condition $c_p$, which is then added to the noise latent before being input into the denoising UNet,as described in \cite{animateanyone}.

\noindent\textbf{Training.} With the design of above, the loss term of this stage is computed as:
\begin{equation}
    \mathcal{L}_{1} = \mathbb{E}_{z_0, c_p, c_r, I_{ref}, \epsilon\sim\mathcal{N}(0,1),t} [||\epsilon - \epsilon_\theta(z_t,c_p, c_r, I_{ref},t)||_2^2],
\end{equation}
where $\epsilon_\theta$ denotes the trainable parameters of the denoising UNet and $t$ is the timestep embedding.

\subsection{Seamless Human Parts Integration.}Another issue is that directly pasting back the rectified human parts $r_{part}$ introduces copy-and-paste artifacts in the edited region, making the generated image appear unnatural. To address this issue, we repaint the area between the background and the rectified human parts, seamlessly integrating them into the target region for a more natural appearance.
 \label{stage-2} 
 
\noindent\textbf{Data Preparation.} Given an image containing human parts like the face or hands, we first locate and crop the target regions and obtain the binary masks using the same approach mentioned in the first stage. For each part, we dilate its binary mask $m$ using the kernel \(k_d\) to obtain the dilated mask \(m_d = \text{dilate}(m, k_d)\). Additionally, we erode the  binary mask \(m\) with a small kernel \(k_e\) to obtain the eroded mask \(m_e = \text{erode}(m, k_e)\). Using the eroded mask, we extract the eroded human part and paste it back into the corresponding region. The erosion process is crucial because the rectified human parts generated in the first stage often exhibit inharmonious edges, which significantly affect the repainting results. By eroding the human part regions, we aim to equip the model with the ability to complete human part edges during the repainting process. This approach helps mitigate issues caused by inharmonious edges, resulting in a more natural and seamless integration of the rectified human parts into the target regions. Suppose the local human part image is denoted as $I$. The corresponding masked image and binary mask can be formulated with Eq.~\ref{eq:inpaint-img} and Eq.~\ref{eq:inpaint-mask}.

\begin{equation}
     I_{f} = I \odot (1-m_d) + I \odot m_e,
     \label{eq:inpaint-img}
\end{equation}
\begin{equation}
     m_{f} = m_d - m_e.
     \label{eq:inpaint-mask}
\end{equation}

Our goal is to predict the area where the binary mask \(m_{f}\) equals one while keeping the other areas unchanged, resulting in the final output $I^{'}$. To achieve this, we first encode the masked image \(I_{f}\) to obtain the masked latent \(l_m = \mathcal{E}(I_{f})\). Next, we downsample the binary mask \(m_{f}\) to match the size of the masked latent \(l_m\). Similar to SD-inpainting, we add five additional input channels for the UNet: four for the encoded masked image \(l_m\) and one for the mask \(m_{f}\). Additionally, we initialize the model with SD-inpainting weights. With this design, the loss term for this stage is computed as follows:
\begin{equation}
\small
    \mathcal{L}_{2} = \mathbb{E}_{z_0, l_m, m_{f}, \epsilon \sim \mathcal{N}(0,1), t} \left[ \left\| \epsilon - \epsilon_\theta(z_t, l_m, m_{f}, t) \right\|_2^2 \right],
\end{equation}
where \(t\) is the timestep embedding.

During inference, we paste the rectified human part \(r_{part}\) back into the corresponding region and predict the unknown area to ensure harmonious integration of the rectified human part. The formulation of \(I_{f}\) during the inference process is given by Eq.~\ref{eq:inpaint-inference}:
\begin{equation}
     I_{f} = I \odot (1-m_d) + r_{part} \odot m_e.
     \label{eq:inpaint-inference}
\end{equation}

% Negative Prompting 
\section{Experiments}
In this section, we begin by detailing the implementation aspects of our approach, followed by a description of the datasets and evaluation protocols used. Additionally, We  present comparative experiments to benchmark our method against previous work, and conduct ablation studies to assess the efficacy of each component in our framework.
% we discuss the potential for further applications of our method. 

Our RealisHuman is trained in two stages: realistic human parts generation and seamlessly integrated the human parts. All experiments are conducted on 8 NVIDIA A800 GPUs. In the first stage, both the main UNet and the Part Detail Encoder are initialized from Real Vision v5.1, and all components are optimizable except for DINOv2\cite{oquab2023dinov2} and VAE encoder/decoder\cite{auto-encoder}. Training is conducted for 50,000 steps with a batch size of 5. In the second stage, only the Inpainting U-Net is optimizable, which is initialized from SD-inpainting\cite{stable_diffusion}. We train the Inpainting U-Net for 20,000 steps with a batch size of 16. For both two stages, the learning rate is set to 5e-5.  The image is resize to a resolution of 512×512. The zero-SNR\cite{zero-snr} and classifier-free guidance(CFG) \cite{cfg} are enabled. The unconditional drop rate is set to 1e-2. We employ HaMeR\cite{hamer} and 3DDFAv3\cite{3ddfa_v3} to estimate the meshes for each human part.
During inference, we adopt a DDIM sampler for 20 denoising steps. We set the hyper-parameter $g_d$ to 5 and $g_e$ to 0.05 times the perimeter of the mask. The images demonstrated in our paper are generated by SDXL\cite{sdxl} and SDXL-LEOSAM\footnote{https://civitai.com/models/43977/leosams-helloworld-xl}.

\subsection{Datasets and Evaluation Protocol.}
We have collected a dataset comprising approximately 58,000 high-quality local hand images and 38,000 high-quality local face images for training our model. To demonstrate the effectiveness of our approach for refining malformed parts, we evaluate its performance on 
%two datasets: UBC  Fashion\cite{zablotskaia2019-ubc} and TED-talks\cite{siarohin2021motion-tedtalk}.
UBC Fashion\cite{zablotskaia2019-ubc} dataset. The human subjects in UBC Fashion exhibit clearly visible hands and faces.
% The human subjects in these datasets exhibit clearly visible hands and faces. 
UBC Fashion consists of 500 training and 100 testing videos, each containing roughly 350 frames. 
%TED-talks includes 1,203 video clips extracted from TED-talk videos on YouTube,
We follow the official train/test split for both UBC Fashion.% and TED-talks. 
Specifically, we use Fréchet Inception Distance (FID)\cite{FID} and the keypoint detection confidence scores of a hand detector or face detector\cite{lugaresi2019mediapipe, zhang2020mediapipe-hand} to evaluate the plausibility of the generated human parts.

\subsection{Results and Comparisons.}
We generate human images with pose guidance on the UBC Fashion dataset
% and TED-talks datasets
using the most advanced human synthesis methods\cite{animateanyone, xu2024magicanimate, zhu2024champ}.  After generating the human images, we located and cropped the regions containing human parts and applied our RealisHuman framework to refine the malformed parts. To mitigate the influence of the relatively small size of human parts in the original images and to better evaluate the metrics, we focused the evaluation specifically on the regions containing human parts.

\begin{table}[htbp]
\centering

%{$\dag$} denotes that the model weights were trained exclusively on the corresponding single data source.}
% \resizebox{\columnwidth}{!}{%
% \begin{tabular}{l|cc|cc}
% \hline
% \textbf{Method} & \multicolumn{2}{c|}{\textbf{UBC Fashion}} & \multicolumn{2}{c}{\textbf{TED-talks}} \\ \hline
%  & \textbf{FID $\downarrow$} & \textbf{Det. Conf. $\uparrow$} & \textbf{FID $\downarrow$} & \textbf{Det. Conf. $\uparrow$} \\ \hline
% AnimateAnyone\cite{animateanyone} & 14.26 & 0.86 & 24.62 & 0.82 \\  
% AnimateAnyone+Ours & 13.02 & 0.91 & 23.54 & 0.90 \\ \hline
% Champ\cite{zhu2024champ} & 27.28 & 0.87 & 37.13 & 0.85 \\  
% Champ+Ours & 25.58 & 0.92 & 35.76 & 0.92 \\ \hline
% MagicAnimate\cite{xu2024magicanimate} & 57.73 & 0.90 & 46.23 & 0.87 \\  
% MagicAnimate+Ours & 55.18 & 0.94 & 45.12 & 0.92 \\ \hline
% % DreamPose\cite{karras2023dreampose}{$\dag$} & -- & -- & -- & -- \\  
% % DreamPose+Ours{$\dag$} & -- & -- & -- & -- \\ \hline
% \end{tabular}
% }
% \label{tab:tab1}
% \end{table}

%{$\dag$} denotes that the model weights were trained exclusively on the corresponding single data source.}
\resizebox{\columnwidth}{!}{%
\begin{tabular}{l|cc|cc}
\hline
\textbf{Method} & \multicolumn{2}{c|}{\textbf{Hand}} & \multicolumn{2}{c}{\textbf{Face}} \\ \hline
 & \textbf{FID $\downarrow$} & \textbf{Det. Conf. $\uparrow$} & \textbf{FID $\downarrow$} & \textbf{Det. Conf. $\uparrow$} \\ \hline
AnimateAnyone\cite{animateanyone} & 14.26 & 0.86 & 20.55 & 0.82 \\  
AnimateAnyone+Ours & 13.02 & 0.91 & 15.44 & 0.90 \\ \hline
Champ\cite{zhu2024champ} & 27.28 & 0.87 & 20.11 & 0.85 \\  
Champ+Ours & 25.58 & 0.92 & 16.74 & 0.92 \\ \hline
MagicAnimate\cite{xu2024magicanimate} & 57.73 & 0.90 &43.12  & 0.87 \\  
MagicAnimate+Ours & 55.18 & 0.94 & 38.81 & 0.92 \\ \hline
% DreamPose\cite{karras2023dreampose}{$\dag$} & -- & -- & -- & -- \\  
% DreamPose+Ours{$\dag$} & -- & -- & -- & -- \\ \hline
\end{tabular}
}
\caption{Comparison of FID and Det. Conf. scores before and after using our method.}
\label{tab:tab1}
\end{table}

\begin{figure*}[h!]
    \centering
\includegraphics[width=1.85\columnwidth]{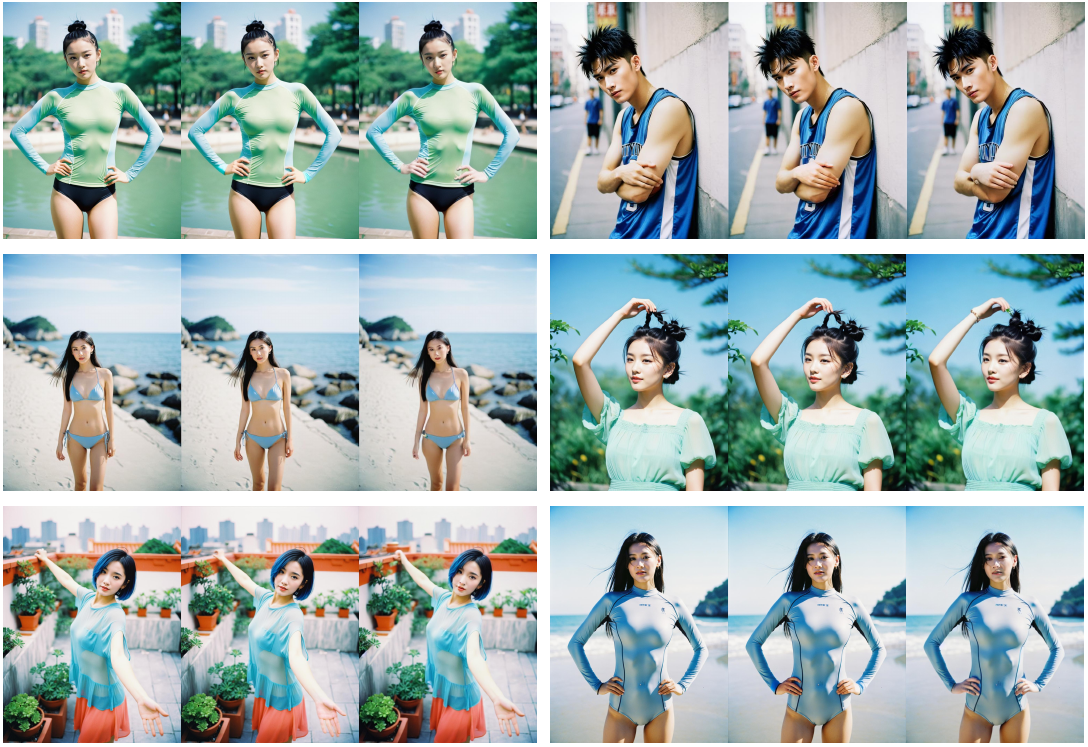}
\caption{Comparison of hand refinement results. Each set of images displays, from left to right, the original image, our method's repair result, and the HandRefiner method's repair result. }
\label{fig:compare-with-handrefiner}
\end{figure*}
In Tab.\ref{tab:tab1}, we report the FID and Det. Conf. scores before and after using our RealisHuman for both face and hand regions. The results demonstrate the effectiveness of our method. Specifically, we observe significant improvements in both metrics after applying our refinement process. The reduction in FID scores indicates that the refined images are perceptually closer to real images, showcasing enhanced realism. Similarly, the increase in Det. Conf. scores reflects improved detection confidence by the detectors, highlighting the structural accuracy and plausibility of the refined face and hand regions.

% In Figure~\ref{fig:4x2grid}, we present the results of our hand image refinement method applied to the UBC Fashion and TED-talks datasets. The images in each row show the results of our refinement method. For each composite image, the left side depicts the hand images before refinement, and the right side illustrates the hand images after applying our method. This comparison highlights the effectiveness of our approach in enhancing both visual quality and structural accuracy.

To evaluate the effectiveness of our method in refining hand images, we compare our method with the popular malformed hands refining method HandRefiner in Fig.\ref{fig:compare-with-handrefiner}. Additionally, we conduct a detailed analysis to illustrate the advantages of our approach.  As shown in Fig.\ref{fig:compare-with-handrefiner}, each comparison figure consists of three horizontally aligned images: from left to right, they display the original image, our method's repair result, and the HandRefiner method's repair result. This figure presents a comprehensive comparison between our method and the HandRefiner method across several critical aspects:
\textbf{(a) Preservation of Hand Details: }Our method excels at maintaining and matching the original details, such as the skin tone of the hands. It demonstrates superior consistency in preserving intricate details, accurately restoring textures and fine features of the hands. As a result, the repaired hands have a more natural and realistic appearance.
%\textbf{(b) Rationality of Hand Structure Repair:} Our approach ensures that the repaired hand structure appears natural and coherent, regardless of the hand's pose.
\textbf{(b)  Effectiveness in Small Hand Repair:} Compare to HandRefiner, our method is particularly effective in repairing smaller hands, meticulously restoring their details and shapes.
% \textbf{(d) Missed Repairs by HandRefiner:} In instances where the HandRefiner method may miss certain repairs, our method provides a comprehensive solution, ensuring no part is left unaddressed.
\textbf{(c) Preservation of Other Regions:} Unlike HandRefiner, which can cause distortions in other areas such as the face while repairing hands, our method preserves the overall integrity and appearance of the image, ensuring that other regions remain unaffected. Effectively showcases these advantages, highlighting the superior performance of our method in hand repair tasks compared to the HandRefiner method. This comparison underscores the efficacy and reliability of our approach in producing high-quality hand restorations.
\begin{figure*}[h!]
    \centering
    \includegraphics[width=1.9\columnwidth]{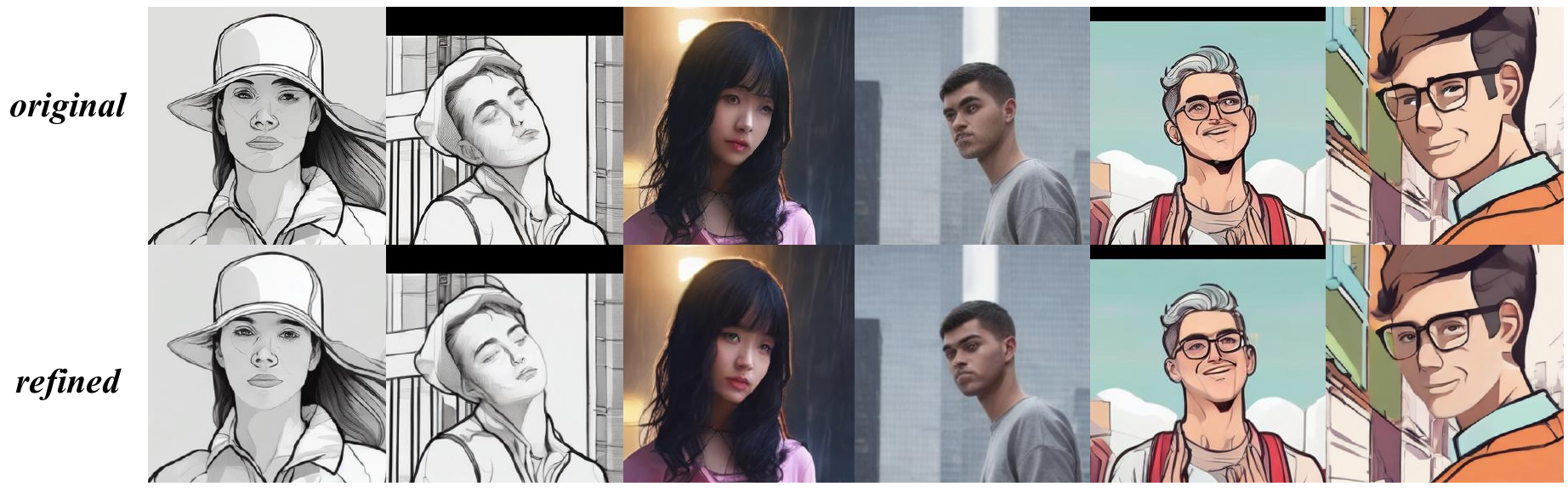} 
    \caption{Comparison of face refinement results. The first row shows the original images, and the second row shows the images after face refinement.}

    \label{fig:face_refine}
\end{figure*}

Additionally, we demonstrate the capability of RealisHuman in facial refinement. As shown in Fig.\ref{fig:face_refine}, our method effectively addresses issues such as distorted facial features and unfocused eyes in the original images, highlighting the efficacy of our approach. The results illustrate that RealisHuman can significantly enhance the realism and accuracy of facial features, further validating the robustness, versatility, and strong generalization capability of our method across various styles of human image restoration. For additional examples and details, please refer to the supplementary materials.
% \subsection{Application.}
% In addition to human parts refinement, our method is capable of addressing structurally fixed local refinement tasks, such as logo refinement, and similar applications. Using face refinement as an example, we first locate and crop the face region and then apply our method to generate realistic faces. These refined faces are seamlessly integrated back into the original images. Our method has achieved remarkably impressive results in face refinement. The results, illustrated in Fig.\ref{fig:face_refine}, demonstrate the effectiveness of our method across various styles of image refinement.

\subsection{Abalation Study.}

\textbf{Effect of the second stage.} 
As discussed above, we address the issue of copy-and-paste artifacts by repainting the transition area between the background and the rectified human parts, ensuring seamless integration into the target region for a more natural appearance. Fig.\ref{fig:copy-and-paste artifacts} compares the results of directly pasting the rectified human part $r_{part}$ with our method. It can be observed that our approach effectively integrates the rectified human parts into the surrounding area without introducing copy-and-paste artifacts.

\begin{figure}[h!]
    \centering
    \includegraphics[width=0.9\columnwidth]{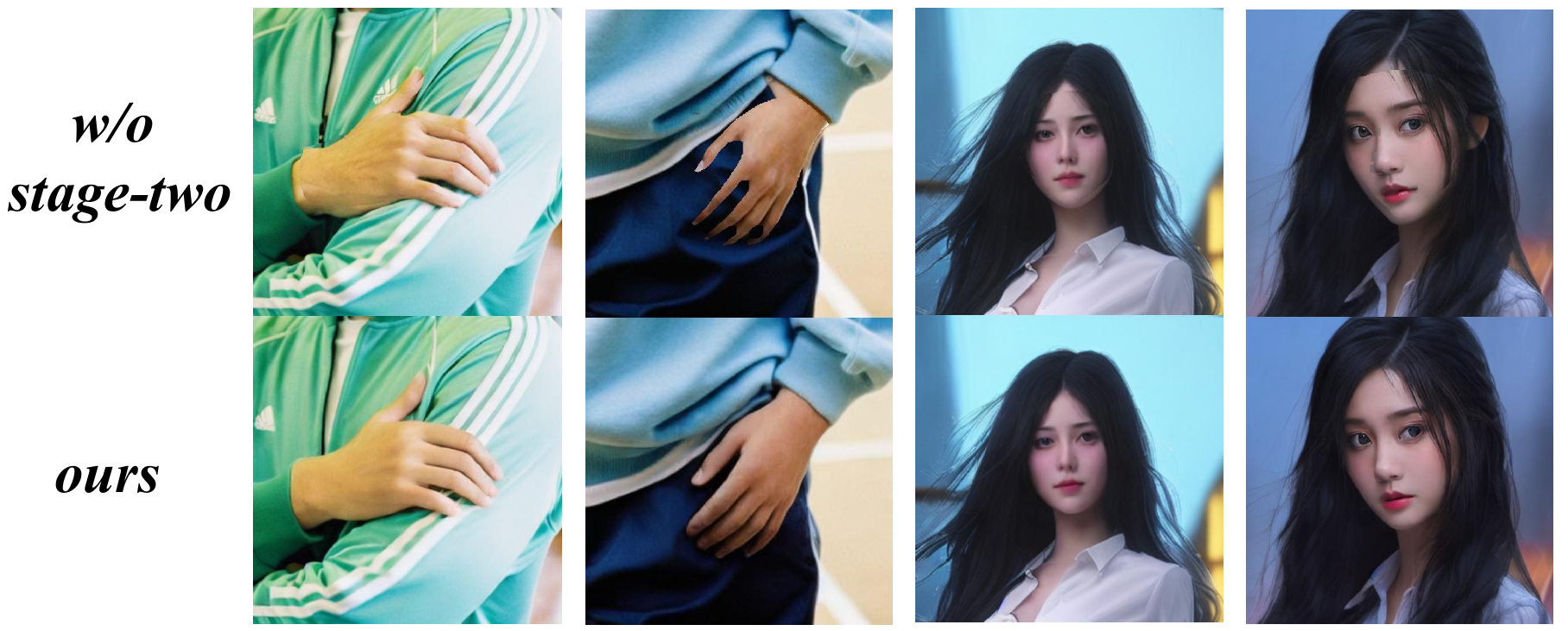} 
    \caption{Comparison of directly pasting the rectified human parts versus our method. The first row shows the results of direct pasting, which exhibit visible artifacts. The second row demonstrates the effectiveness of our method in achieving seamless integration.}
    \label{fig:copy-and-paste artifacts}
\end{figure}
\noindent\textbf{Effect of the eroded mask $m_e$.} As discussed above, the eroded mask $m_e$ is used to mitigate the effects of inharmonious edges in the second stage. Without the eroded mask, these inharmonious edges can hinder the seamless integration of the rectified human parts with their surroundings, leading to the generation of discordant elements such as hair, watches, and other artifacts. We illustrate the impact of the eroded mask in Fig.\ref{fig:copy-and-paste artifacts}, comparing images processed with and without it. The first row shows the results without the eroded mask, where noticeable artifacts are present. The second row demonstrates the results when using the eroded mask, which effectively reduces edge artifacts and achieves a smoother integration.
\begin{figure}[h!]
    \centering
    \includegraphics[width=0.9\columnwidth]{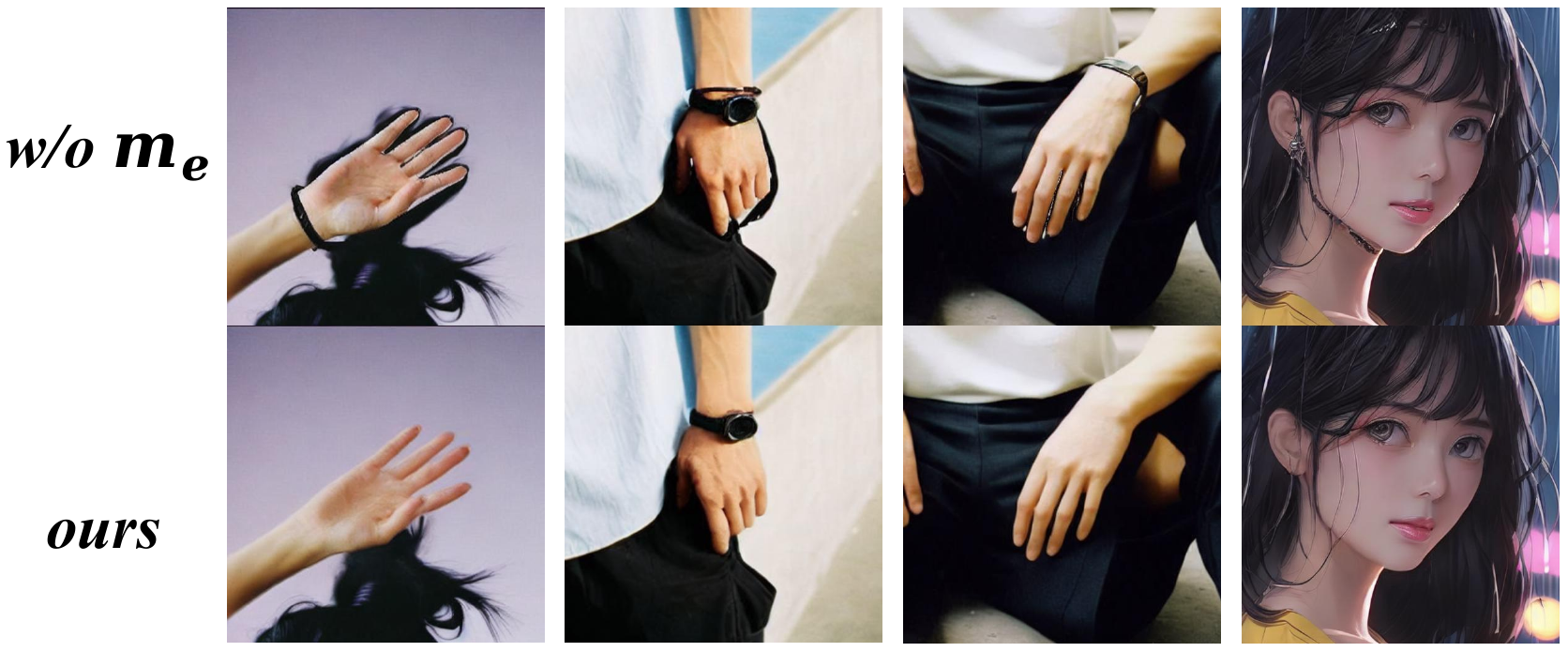} 
    \caption{Comparison of results with and without the eroded mask $m_e$. The first row shows visible edge artifacts without the eroded mask. The second row demonstrates improved integration using the eroded mask, reducing edge artifacts.}
    \label{fig:copy-and-paste artifacts}
\end{figure}

\section{Limitations and Discussion}
\begin{figure}[h!]
    \centering
    \includegraphics[width=0.8\columnwidth]{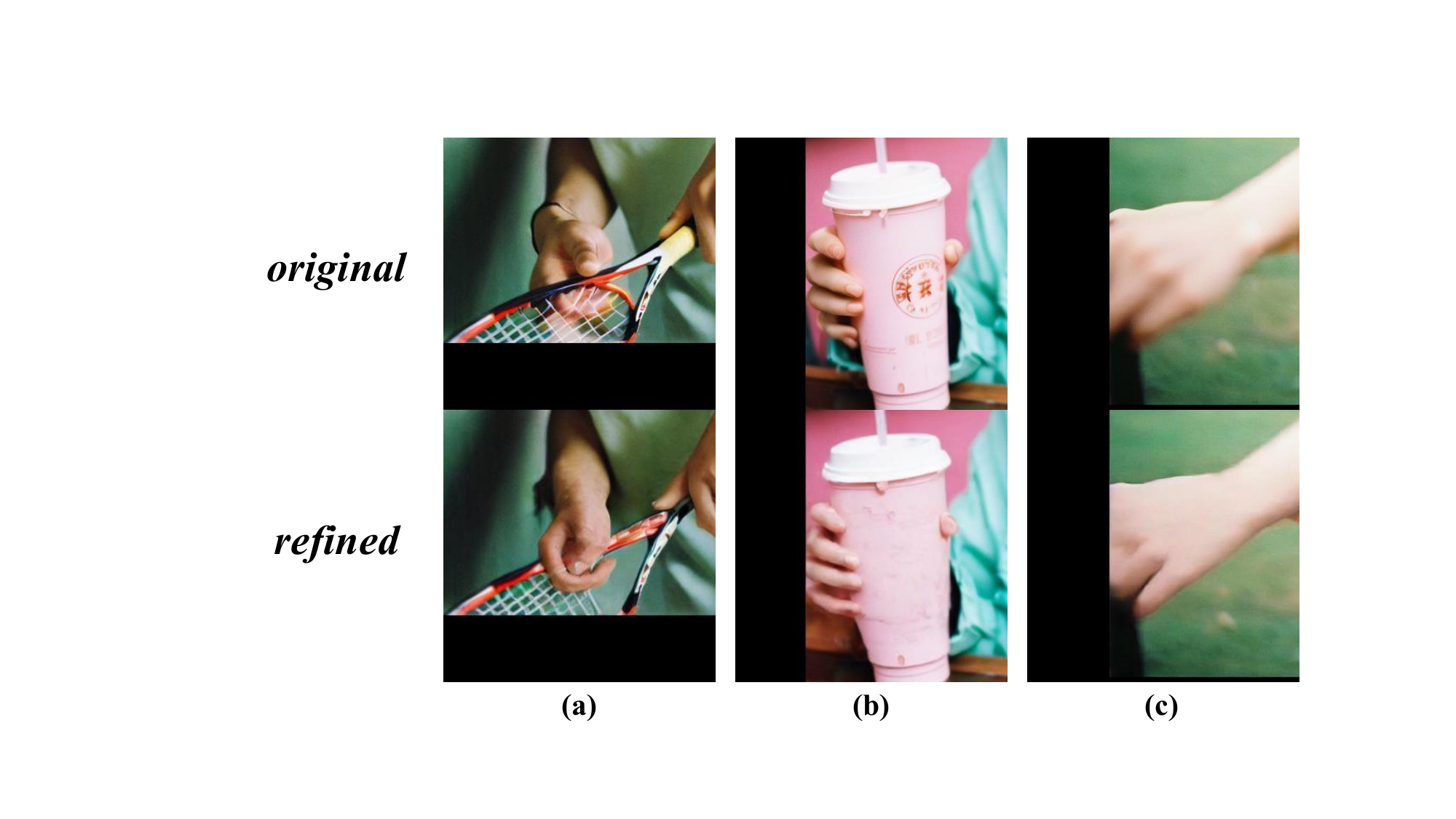} 
    \caption{Illustration of the limitations of our method. }
    \label{fig:limitation}
\end{figure}

While our method has demonstrated notable improvements in refining and reconstructing human hands, it still faces several challenges, as illustrated in Fig.\ref{fig:limitation}
. Firstly, the method may struggle to accurately reconstruct interaction between hands and objects. Secondly, it may fail to maintain consistency in the presence of objects. Thirdly, when the original hand is severely distorted, the method may be unable to estimate the correct hand pose, leading to unsuccessful hand reconstruction. Addressing these issues will be the focus of our future work, potentially incorporating more sophisticated modeling techniques or leveraging additional contextual information to improve performance in these areas.
\section{Conclusion}
In this paper, we introduced RealisHuman, a novel post-processing solution for refining malformed human parts in generated images. Our method operates in two stages: first, generating realistic human parts using the original malformed human parts as the reference to maintain consistent details; second, seamlessly integrating the rectified human parts by repainting the surrounding areas. This framework effectively addresses the challenges of human parts generation and can be extended to other local refinement tasks, such as logo refinement. Comprehensive experiments demonstrate significant improvements in both qualitative and quantitative measures, validating the effectiveness and robustness of our approach.

\bibliography{aaai25}

\end{document}